\documentclass{sig-alternate}

\usepackage[hyphenbreaks]{breakurl}
\usepackage[hyphens]{url}
\usepackage{amsmath} 
\usepackage{listings}
\usepackage{fancyvrb}

\permission{}
\copyrightetc{}
\permission{Copyright is held by the International World Wide Web Conference Committee (IW3C2). IW3C2 reserves the right to provide a hyperlink to the author's site if the Material is used in electronic media.}
\conferenceinfo{WWW'14,}{April 7--11, 2014, Seoul, Korea.} 
\copyrightetc{ACM \the\acmcopyr}
\crdata{978-1-4503-2744-2/14/04. \\ http://dx.doi.org/10.1145/2567948.2578846 }

\begin{document}
 
\title{A Semantic Web of Know-How: \\Linked Data for Community-Centric Tasks}

\numberofauthors{3} 
\author{
\alignauthor
Paolo Pareti\\
       \affaddr{Edinburgh University}\\
       \email{p.pareti@sms.ed.ac.uk}
\alignauthor
Ewan Klein\\
       \affaddr{Edinburgh University}\\
       \email{ewan@inf.ed.ac.uk}
\alignauthor 
Adam Barker\\
       \affaddr{University of St Andrews}\\
       \email{adam.barker@st-andrews.ac.uk}
}

\maketitle
\begin{abstract}
\noindent This paper proposes a novel framework for representing community `know-how' on the Semantic Web. Procedural knowledge generated by web communities typically takes the form of natural language instructions or videos and is largely unstructured. The absence of semantic structure impedes the deployment of many useful applications, in particular the ability to discover and integrate know-how automatically. We discuss the characteristics of community know-how and argue that existing knowledge representation frameworks fail to represent it adequately. We present a novel framework for representing the semantic structure of community know-how and demonstrate the feasibility of our approach by providing a concrete implementation which includes a method for automatically acquiring procedural knowledge for real-world tasks.
\end{abstract}


\section{Introduction}
\noindent Know-how, or procedural knowledge, is an important resource that communities on the Web can create, share and benefit from. Representing this resource semantically, however, is still an open problem. This paper addresses this problem proposing a novel knowledge representation framework that can effectively represent community know-how. The main goals of this paper are (1) to motivate the need for a semantic representation of know-how, (2) to describe the features of the proposed framework and (3) to demonstrate its feasibility.
\par Two motivations for this work are the importance of know-how as a community resource and the benefits that can accrue from representing it semantically. As the amount of know-how on the Web grows, it becomes more and more common to look for instructions or solutions to everyday problems on the Web. However, unlike declarative knowledge, procedural knowledge is not yet adequately addressed by the Semantic Web \cite{Thimm2012}. Without a shared semantic representation of know-how, discovering and integrating a large number of procedures remains a difficult challenge. This is an issue for many common tasks, like the organisation of an academic event, which involve collecting and sharing a large amount of know-how and using several tools and resources. In general, this is a limitation to the potential of the Web in empowering user communities with the vast amount of existing know-how, services, and efficient communication mechanisms.
\par In order for this potential to be fully exploited it is not enough just to make these resources available on the Web. User communities need intelligent systems to leverage the large amount of web resources and achieve the tasks they are interested in. The combination of the Social Web with the Semantic Web is at the basis of what Tom Gruber has called \emph{collective-knowledge systems} \cite{Gruber2008CollectiveIntelligence}. A semantic representation of procedures would allow the development of those systems by bringing similar benefits to those that drive the Semantic Web vision. These benefits are not limited to the automatic discovery and integration of procedures. Several other applications have been suggested, like in Activity Recognition \cite{Perkowitz2004MiningModels} and Robotics \cite{Tenorth2011WebEnabled}. 
\par Many knowledge representation frameworks have already been created to represent procedures semantically. These frameworks find applications, for example, in Automated Planning, Workflow Management and Web Services. In all of those frameworks human participation is more or less limited. A large subset of real-world tasks, however, involve not just individual human actors. Instead, communities have an important role both in decision making and in the execution of these tasks. Such community centric tasks cannot be represented efficiently with existing knowledge representation frameworks due to the following difficulties:
\begin{itemize}
        \item Uncertainty and knowledge gaps.
        \item Lack of a centralised knowledge base.
        \item Evolution over time.
        \item Extensive community participation.
\end{itemize}
\noindent The main contribution of this paper is a novel knowledge representation framework that overcomes those difficulties thanks to the following characteristics:
\begin{itemize}
        \item It provides a lightweight representation of generic tasks as Linked Data which can be shared on the Semantic Web. 
        \item It allows know-how to be represented in a distributed fashion across different knowledge bases. All the information relevant to a task can be retrieved dynamically and integrated automatically.
        \item It has a scalable approach to reasoning. The computational effort to answer a query is distributed between the different SPARQL endpoints\footnote{\burl{http://www.w3.org/TR/sparql11-overview/}} of the various knowledge bases.
        \item Communities are seen as the main producers and consumers of know-how. This means that collaboration between community members is not limited to the completion of tasks, but it also takes form of collaborative creation and sharing of know-how.
\end{itemize}

\section{Community centric tasks} \label{definitioncommunitycentrictasks}
\noindent The proposed definition of community centric tasks is better thought not as a crisp concept, but instead as a set of properties which can determine how strong is the human and social aspect of a given task. At the opposite ends of the spectrum we find strongly community centric tasks, like collaborating to the development of a project, and tasks which are not community centric, like counting all the integers from 1 to 10. In the middle of the spectrum it is possible to find borderline cases that share some, but not all, of the human and social aspects. Cooking a pancake is an example of a borderline case. Examining real-world examples, we have identified six properties that define community centric tasks:
\newtheorem{property}{Property}
\newcommand{\mypropertydef}[1]{\begin{property}#1\end{property}}
\begin{property} \label{propertycollaborative} \normalfont{\textbf{Collaborative}. The task requires collaboration between different agents.}\end{property}
\begin{property} \label{propertycommunityinterest} \normalfont{\textbf{Of general interest within a community}. There is an interest in sharing knowledge and other resources about the task.}\end{property}
\begin{property} \label{propertydistributedknowledge} \normalfont{\textbf{Distributed knowledge}. Knowledge on how to perform the task is distributed across several knowledge sources. These sources can be digital web resources as well as the human knowledge of the members of a community.}\end{property}
\begin{property} \label{propertyvarietyapproaches} \normalfont{\textbf{Variety of approaches}. There are a potentially large number of ways of performing the task.}\end{property}
\begin{property} \label{propertygranularityrange} \normalfont{\textbf{Wide range of granularity}. The steps required to perform the task have different levels of abstraction. Typically, the most vague and abstract steps are solved better by humans. The most specific and concrete steps might be automated by machines.}\end{property}
\begin{property} \label{propertyknowledgegaps} \normalfont{\textbf{Knowledge gaps}. It is not possible to know all the details of the task \textrm{a priori}. Some of the required know-how might be available elsewhere (i.e.\ not yet discovered) or it might not even exist. An immediate consequence of this property is that it is not possible to plan all the details of the task a priori. Another important consequence is that acquiring the missing knowledge is an integral part of the process of performing a task.}\end{property}

\section{An example scenario}
\noindent This section introduces the example scenario that will be used in the reminder of the paper. This example will illustrate the various aspects of community centric tasks. A discussion will then highlight the main applications of the proposed framework, namely: knowledge discovery, knowledge creation, execution and collaboration.
\par In this example a member of the academic staff is attempting to organise a conference. Being the first time she performs this task, she does not have all of the necessary know-how from the start. For this reason, she is interested in benefiting from all the resources available on the Web which are relevant to this task. She is also interested in collaborating with multiple persons to share the organisational effort (Property \ref{propertycollaborative}). There are different ways to organise a conference (Property \ref{propertyvarietyapproaches}) and useful know-how is distributed between multiple persons and resources (Property \ref{propertydistributedknowledge}). This task involves both abstract steps, like deciding the topic of a workshop, and concrete ones, like adding the program schedule on the conference website (Property \ref{propertygranularityrange}). As all conferences are different, the task of organising one will ultimately involve solving unique challenges for which no solution already exists (Property \ref{propertyknowledgegaps}). Being a common task in the academic community (among others), there is an interest in sharing relevant resources such as instructions, tools and past experiences (Property \ref{propertycommunityinterest}).

\subsection{Knowledge discovery} \label{subsec:knowledgediscovery}
\noindent In this scenario, the first problem faced by the organiser of the conference is the discovery of all the relevant know-how. One could argue that the current web search engines can efficiently retrieve the best results for a given query. For example, a query like ``How to organise a conference?" will most likely retrieve a set of instructions on how to organise the conference. This set of instructions might also have some hyperlinks to additional resources, like tools to calculate the conference budget.
However, as many different solutions to this problem exist (Property \ref{propertyvarietyapproaches}), the organiser of the conference might want to answer more complex questions. For example, she might want to know ``What are the other ways to organise a conference?", ``Who has already followed those instructions?" or ``What are the other tools to calculate the conference budget?". Answering these question using the existing search engines might not be possible, or it might require looking at all the search results extensively.
\par The first step to answer those queries automatically is to be able to identify a process or any of its parts with precision across distributed knowledge bases (Property \ref{propertydistributedknowledge}). This could be achieved by associating these concepts with URIs. If the user then discovers one such URI, she could search for any other resource that is related with that URI. In order to do so, relations between URIs should be represented using RDF\footnote{\burl{http://www.w3.org/TR/rdf-primer/}} and be available on the Semantic Web. This search can then be automated using the SPARQL query language over a set of knowledge bases, or using a Semantic Web index like Sindice.\footnote{\burl{http://sindice.com/}}
\par Once the organiser of the conference has found a URI that corresponds to the concept of ``organising a conference", she can discover all the related entities. Following the relations with meaning ``has method", for example, she can find the URIs of the other methods to organise a conference. From the URI of an alternative method she can then retrieve further resources, like the document where the instructions for that method are written. A different relation, for example ``has requirement", can lead to the discovery of other facts, like the things to check before starting the procedure.
\par The organiser of the conference uses this method to find the URIs of five different conference budget management tools (and consequently their website). How can she be sure that there is no other tool available? This cannot be guaranteed to be true since it is impossible to assume that all the required knowledge to perform a task already exists (Property \ref{propertyknowledgegaps}). To compensate this necessary uncertainty, the proposed framework allows the creation of the missing knowledge even after the execution of the task has already started.

\subsection{Knowledge creation}
\noindent Know-how which is found to be missing can be produced both by automatic knowledge extraction systems and by community efforts. For example, the organiser of the conference does not find specific instructions on how to organise the catering service. In order to fill this knowledge gap she decides to ask for advice to a senior colleague of hers who has already organised a similar event in the past. After receiving this advice, she decides to publish it on a web page because it could be useful for other members of the academic community (Property \ref{propertycommunityinterest}). Now she (and anybody else) has the possibility to generate a URI to represent the information contained in that web page and to link it to the URI representing the ``organise catering" step. After this relation is published on the Semantic Web, anybody who is interested in the ``organise catering" process will be able to find this additional information automatically.
\par Similarly, the same process URI can be linked to multiple resources. These resources could be web pages, text snippets of different verbosity, pictures, videos or any other resource type. Different users would then be able to select the most convenient description(s) for their needs. Some users might prefer textual descriptions, others might prefer visual instructions; some might prefer less verbosity, others might need more details. In our example, the organiser of the conference is looking at the step: ``setup the meeting room" but she is confused by the textual description of the various possible setups. Fortunately, somebody has published the relation between that URI and a schematic picture of how the various meeting room setups look like. Thanks to this connection, she does not have to search for this picture, and she can be automatically offered it as a visual aid. The creation of know-how can be seen as a form of collaboration to solve complex problems. In the next section we will see a more direct form of collaboration, multiple agents completing the steps of the same task.

\subsection{Execution and collaboration}
\noindent The execution of a process can involve several collaborators. Being able to refer to different parts of a process using URIs allows collaborators to know exactly which part of the process they should work on. It is also possible to keep track of the progress made in each of those steps to identify, for example, which steps of a procedure still have to be completed. It is important to notice that the proposed framework is intended to \emph{describe} how a process can be done and not to \emph{prescribe} how it should be done. This means that there is no strict requirement to follow all and only the steps of a procedure.
\par The semantic connections between different steps (e.g.\ the order of execution) can help coordinating the collaboration. In the example we are considering, the step $A$: ``organise the catering service" should be done only after steps $B$: ``decide the location for the conference" and $C$: ``decide on a preliminary budget" have been completed. Additionally, these three steps have been delegated to three different collaborators. Having a shared semantic representation of this dependency, the collaborator responsible for step $A$ can be automatically notified (that work on $A$ can be started) as soon as both steps $B$ and $C$ have been completed.
\par This automatic notification system is only one of many possible ways to support community centric tasks. Other ways to support community know-how come from a large number of existing tools and websites. For example, there are resources supporting project management and collaboration (e.g.\ Trello),\footnote{\burl{http://trello.com/}} creation of community generated know-how (e.g.\ wikiHow)\footnote{\burl{http://www.wikihow.com/}} and task automation (e.g.\ IFTTT).\footnote{\burl{http://ifttt.com/}} Without a shared knowledge representation format, however, the knowledge and functionalities offered by those resources remain isolated and cannot be automatically integrated. The proposed framework is not supposed to replace those resources, but instead to allow their integration. For example, the proposed knowledge representation framework would allow the development of a system which could:
\begin{itemize}
\item Help the organiser of the conference to discover a procedure to organise conferences along with its semantic representation.
\item Automatically load this representation into the project management tool of her choice.
\item Automatically generate instructions for task automation tools from the representation of some steps.
\item Give collaborators the freedom to use other tools to visualise and manage the same procedure if they prefer.
\end{itemize}

\section{The proposed framework}
\noindent Before describing the details of how community centric tasks are represented, the three most important features of the proposed framework will be discussed. The first feature is the way it is decentralised on the Semantic Web. Knowledge from several knowledge bases can be integrated by delegating query execution to their respective SPARQL endpoints. This feature is also the main novelty of this approach. The second feature is that this framework is generic and captures concepts like ``process decomposition" and ``dependency" which are applicable in most domains. The third feature is the lack of a rigid logical formalism. This last feature allows the framework to be more robust to missing or erroneous knowledge. However this comes to the cost of a reduced inference power.

\subsection{Comparison with existing frameworks}
\noindent Several ontologies, like OWL-S \cite{Martin2004owls}, can be used to describe processes on the Semantic Web. All of those ontologies, however, use the Semantic Web primarily as a medium to publish and share structured data. This means that the semantic description of a process needs to be retrieved before a system can reason about it. This causes two main disadvantages. The first is the need to adopt an additional technology to reason about the retrieved knowledge. In the case of OWL-S, this could be an OWL reasoner. The second is a limitation in scalability. Since the reasoning process is centralised, all the relevant knowledge about a process needs to be available in a local knowledge base before the reasoning process can begin.
\par In the proposed approach, the reasoning process happens directly on the Semantic Web by means of SPARQL queries.  Questions like: ``What are the steps of process $X$?" or: ``Has step $Y$ been completed?" can be answered by integrating the results of a single SPARQL query over a number of knowledge bases (or Semantic Web indexes). A combination of SPARQL queries can also be used to answer more complex queries like ``Which steps of process $X$ have not been completed?". An advantage of this approach is that SPARQL queries can selectively extract relevant information from distributed knowledge bases without the need to retrieve all the knowledge they contain. The reasoning process can be said to be decentralised as the computational effort required to answer a query is shared between the various SPARQL endpoints. Another advantage is that it does not require the adoption of any technology other than RDF and SPARQL.
\par Other reasons why existing process ontologies do not offer a convenient representation of community centric tasks are domain specificity and logic heaviness. Domain specificity (e.g\ found in Web Services ontologies) can be considered a disadvantage in representing community centric tasks, which are inherently generic. Domain specific ontologies applied to a generic domain can be unnecessary complex and add the risk of representing unintended semantics. The OWL-S ontology, for example, expects a process to be a ``specification of the ways a client may interact with a service" \cite{Martin2004owls}.
\par Logic heavy ontologies (e.g\ those used in Automated Planning) are not compatible with the fact that community centric tasks can be represented across distributed knowledge bases. Having a distributed representation of a process using a logic-heavy ontology might not be possible. For example, dividing a PSL \cite{Gruninger2003PSL} process into several parts might cause them to be locally inconsistent. Similarly, logic heaviness can hinder the integration of different process representations. For example, multiple PDDL \cite{mcdermott98pddl} planning domain descriptions cannot be integrated automatically.
\par The proposed framework, instead, aims to be lightweight and generic. To achieve this, the semantic representation of processes is based on a simple RDFS\footnote{\burl{http://www.w3.org/TR/rdf-schema/}} vocabulary. If required, this vocabulary could be extended with additional terms to accomodate the needs of domain specific applications. The prefixes used in the reminder of the paper can be found in Table \ref{tab:PrefixesUsedInTheDocument}. A complete listing of the core vocabulary is given in Table \ref{tab:vocabulary}. This vocabulary can be divided into a process and an execution ontology. Those ontologies are used, respectively, to describe the semantics of a process and the state of its executions. It should be noted that those ontologies, although minimal, are sufficient to support the core functionalities of the proposed framework.

\begin{table}
        \centering
                \begin{tabular}{ | l | p{6.5cm} |}
                        \hline
                        \bfseries{Prefix} & \bfseries{Namespace} \\ \hline
                        : & \burl{http://example.ex/} \\ \hline
                        rdf: & \burl{http://www.w3.org/1999/02/22-rdf-syntax-ns#} \\ \hline
                        rdfs: & \burl{http://www.w3.org/2000/01/rdf-schema#} \\ \hline
                        prohow: & \burl{http://vocab.inf.ed.ac.uk/prohow#} \\ \hline
                        proex: & \burl{http://vocab.inf.ed.ac.uk/proex/0.1#} \\ \hline
                \end{tabular}
        \caption{Prefixes used in the document}
        \label{tab:PrefixesUsedInTheDocument}
\end{table}

\begin{table}
        \centering
                \begin{tabular}{ | l | p{4.5cm} |}
                        \hline
                        \bfseries{Term} & \bfseries{Definition when $X$ is the subject and $Y$ is the object} \\ \hline
                        \burl{prohow:has_step} & $Y$ can be accomplished as part of $X$  \\  \hline
                        \burl{prohow:requires} & $Y$ should be accomplished before doing $X$ \\ \hline
                        \burl{prohow:has_method} & $Y$ can be accomplished instead of $X$ \\ \hline
                        \burl{proex:has_goal} & The execution $X$ is trying to accomplish $Y$ \\ \hline
                        \burl{proex:succeed_in} & $X$ has been accomplished in execution $Y$ \\ \hline
                        \burl{proex:failed_in} & $X$ has failed in execution $Y$ \\ \hline
                \end{tabular}
        \caption{The vocabulary of the ontology}
        \label{tab:vocabulary}
\end{table}

\subsection{The process ontology}
\noindent As previously discussed in Section \ref{subsec:knowledgediscovery}, the need to identify entities across distributed knowledge bases leads to the adoption of URI identifiers. In its simplest form, this ontology identifies a process as a single URI. For example, the URI \burl{:organise_conference} can be used to represent the process of organising a conference. The connection between this entity and the resources that describe it can be made explicit using the Open Annotation Data Model.\footnote{\burl{http://www.openannotation.org/}} For example, this model can represent the fact that web page \burl{http://www.wikihow.com/Organize-a-Conference} (which contains instructions on how to organise a conference) is related to the URI \burl{:organise_conference}. The annotations created using the Open Annotation Data Model can also target multimedia resources and resource fragments.
\par Processes can be seen at different levels of abstraction. The most high-level steps provide a short summary of what the process involves. The most fine grained steps provide explicit details on how the process can be performed. The relation \burl{prohow:has_step} can be used to connect a process to one of its possible steps. Since that step can have further sub-steps (and so on) arbitrary levels of abstraction are allowed. To explain the use of this relation, let us consider the URI \burl{:choose_conference_venue}, which is used to represent the process of choosing a venue for the conference. This URI can be connected to the text snippet of web page \burl{http://www.wikihow.com/Organize-a-Conference} which describes this step. The following RDF triple (expressed using the Turtle\footnote{\burl{http://www.w3.org/TeamSubmission/turtle/}} syntax) represents the fact that choosing a venue is a step within the process of organising a conference:
\begin{Verbatim}
:organise_conference prohow:has_step
        :choose_conference_venue .
\end{Verbatim}

\noindent The proposed RDFS vocabulary does not provide a distinction between objects, conditions or processes. This distinction, if required, can be introduced as an extension of this vocabulary. To demonstrate why this distinction is not always necessary, let us consider the following example. As part of the organisation process, a preliminary budget (URI \burl{:preliminary_budget}) is required before starting the organisation of the catering service (URI \burl{:organise_catering}). This information could be represented in different ways. For example, the process ``organise the catering service" might require: (1) the object ``preliminary budget" as an input, (2) the precondition ``the preliminary budget is known" to be true or (3) to start after the process ``decide on a preliminary budget" has been completed. These three representations make different assertions on whether the entities are objects, conditions or processes. However they all convey the meaning that there is a dependency between entity ``organise the catering service" and entity ``preliminary budget". An agent which is only interested in expressing this dependency might not want to commit to a specific classification of the entities. To avoid unnecessary semantics, this class distinction is not part of the proposed vocabulary. The concept of dependency can be expressed using the \burl{prohow:requires} relation as follows:
\begin{Verbatim}
:organise_catering prohow:requires 
        :preliminary_budget .
\end{Verbatim}

\noindent The \burl{prohow:has_method} relation can be used to connect a process with the different ways to achieve it. For example, if there is a web page $X$ containing instructions on how to choose a conference venue (URI \burl{:choose_venue_method}), these instructions could be connected to the corresponding step of our example (URI \burl{:choose_conference_venue}). This can be represented as the following triple:
\begin{Verbatim}
:choose_conference_venue prohow:has_method 
        :choose_venue_method .
\end{Verbatim}
\noindent This example shows how different resources can be linked together using this vocabulary. By following the \burl{prohow:has_method} relation the organiser of the conference can discover the URIs of various approaches (e.g.\ \burl{:choose_venue_method}) and then discover the related web resources (e.g.\ web page $X$).

\subsection{The execution ontology}
\noindent When it comes to performing tasks, it is useful to have knowledge not only about the task, but also about its current execution. This is particularly important in case of collaboration, as it allows collaborators to have a shared understanding of the progress made in completing the task. This execution ontology extends the semantic representation of a process with the concept of an execution. An execution can have one or more goals, which are the tasks that the execution is trying to achieve. In our example, the task of organising a conference (URI \burl{:organise_conference}) is the goal of the execution \burl{:execution1}. This can be expressed with the following triple:
\begin{Verbatim}
:execution1 proex:has_goal :organise_conference .
\end{Verbatim}
\noindent Other members of the community who want to follow the same process to organise a different conference will use a different execution URI. Different executions are considered different environments and processes which have been completed in one might not have been completed in another. The \burl{proex:succeeded_in} relation can be used to indicate that a process has been completed within an execution. In our example, the fact that the catering step (URI \burl{:organise_catering}) has been completed can be represented as follows:
\begin{Verbatim}
:organise_catering proex:succeeded_in :execution1 .
\end{Verbatim}
\noindent The \burl{proex:failed_in} relation can be used as the \burl{proex:succeeded_in} relation but to indicate that a process has failed instead. This can be seen as an indication that an alternative approach should be tried.

\section{Feasibility study}
\subsection{Knowledge acquisition experiment}
\noindent A common bottleneck of Semantic Web frameworks is the difficulty in generating semantic data efficiently. This leads to the question of how to produce the semantic representation of a task. The answer to this question varies depending on whether the available description of the task is unstructured or semi-structured.
\par Unstructured know-how is often represented in natural language. Most of the existing approaches for extracting procedural knowledge from natural language texts use Natural Language Processing (NLP) techniques \cite{Addis2011FromUnstructured,Jung2010AutomaticConstruction,Schumacher2012ExtractionOf}. Other approaches have also combined NLP techniques with Machine Learning \cite{Song2011ProceduralKnowledge} and statistical analysis \cite{Fukazawa2010AutomaticModeling}.
\par The main limitation of these knowledge extraction systems is the difficulty in achieving high accuracy. This limitation is due to the lack of structure of the documents to analyse. A large number of know-how resources already available on the web, however, are partially structured. Our knowledge extraction experiment focuses on the semi-structured articles collected by the wikiHow community. This community creates generic know-how collaboratively, as anybody has the possibility to edit and improve an article. The wikiHow website already contains over 160,000 semi-structured procedures. This structure takes form, for example, as an explicit subdivision of processes into steps. Another example of this structure is the enumeration of the requirements of a process, like tools and ingredients. 
\par It is important to note that this structure has been created collaboratively by the whole community. A more structured representation of a process, in fact, is not only more understandable by machines, but first and foremost by people. This structure is considered good practice as it makes the overall description of the process clear and reduces ambiguities.
\par Using this structure, the automatic knowledge extraction system we developed crawled the wikiHow website generating a semantic representation of the articles. For each article, the main task was identified along with its requirements, steps and substeps (if present). A subset of the extracted knowledge is now available on the Semantic Web.\footnote{SPARQL endpoint: \burl{http://dydra.com/paolo-pareti/knowhow6/sparql}} 

\subsection{Knowledge discovery and execution}
\noindent The large volume of data extracted by this knowledge acquisition experiment allowed us to test a concrete implementation of our framework with real-world data.\footnote{This implementation can be found at: \burl{http://bitbucket.org/paolopareti/know-how-explorer}} In order to support distributed knowledge-bases, this system can retrieve knowledge from an arbitrary number of SPARQL endpoints. Two functionalities are currently allowed: knowledge discovery and process execution.
\par The first step of knowledge discovery is the identification of the URI of an entity of interest. This entity is not necessarily a procedure, but it can also represent any of its parts, like a step or an ingredient. At the moment, the discovery of this entity is done using a keyword search. This search generates a SPARQL query that is performed over all the SPARQL endpoints. The results are collected and presented to the user which can decide which one to select.
\par Once the user has selected a URI, the system can query the different SPARQL endpoints to retrieve any related entity. If the URI corresponds to a task, related entities might be requirements, steps, or the more high-level tasks it is part of. If it corresponds to an object, these entities might be the processes that use or produce that object. The user can select one of those entities and continue exploring.
\par If the user wishes to accomplish one of the discovered tasks, the system can publish the information about this activity on the Semantic Web. Other users can then join this activity and see the progress that has been done. If they decide to collaborate, for example performing one of the steps, they can publish this information too. The next user that visualises that step will be able to see that it has been completed.

\section{Conclusion}
\noindent This paper proposed a novel framework to semantically represent community know-how and demonstrated its feasibility. The concept of ``community centric tasks" was introduced to describe the particular properties of this type of knowledge, like strong community participation and lack of a centralised knowledge base. Considering those properties, existing knowledge representation frameworks were not found to be effective at representing this type of tasks.
\par We have discussed the novelty of the proposed framework and why it provides a better representation of community centric tasks. In particular, this framework supports decentralised reasoning and it allows automatic retrieval and integration of know-how from distributed knowledge bases. Another novelty is that it is entirely built on top of Semantic Web standards. The semantic representation of processes is based on a simple RDFS vocabulary. 
\par To demonstrate that those properties are effective at representing community know-how, we have described a feasibility study based on real world data. Our knowledge acquisition experiment showed the feasibility of automatically extracting the semantic representation of procedures from semi-structured descriptions. Although this approach is not applicable where no structure is available, we argue that this is not a significant limitation. In fact, a large volume of semi-structured know-how is already available on various websites like wikiHow. Many communities interested in know-how are spontaneously adopting more structured descriptions of procedures to make them more understandable.
\par Lastly, we have presented a concrete implementation of our framework that uses the extracted knowledge. This system allows users to explore know-how which is automatically integrated from distributed knowledge bases. Users can also publish information about the activities they are performing and share them with collaborators.

\newpage

\bibliographystyle{abbrv}
\bibliography{phdbib} 

\balancecolumns

\end{document}